\documentclass[letterpaper, 10 pt, conference]{ieeeconf-modified}
\IEEEoverridecommandlockouts                              

\usepackage{graphicx}

\usepackage[table,xcdraw]{xcolor}

\usepackage{times}
\usepackage{amsmath} 
\usepackage{amssymb}
\usepackage{mathtools}

\usepackage[numbers]{natbib}
\usepackage{textcomp}
\usepackage{array}
\usepackage{multirow}
\usepackage{booktabs}

\usepackage{hyperref}
\usepackage[capitalize]{cleveref}

\usepackage{algorithm}
\usepackage[noend]{algpseudocode}

\crefname{equation}{}{}  

\newcommand{\gitlink}{https://dlr-alr.github.io/2022-iros-planning}

\DeclareMathOperator*{\argmin}{argmin}

\newcommand{\Feasibility}{\phi}
\newcommand{\qconfig}{q}
\newcommand{\QPath}{Q}
\newcommand{\QPathNet}{Q_\mathrm{p}}

\newcommand{\Frames}{F}

\newcommand{\funkin}{f}
\newcommand{\funclip}{c}
\newcommand{\distfield}{D}

\newcommand{\OMPCost}{U}
\newcommand{\OMPCostLength}{U_\text{L}}
\newcommand{\OMPCostCollision}{U_\text{C}}
\newcommand{\OMPCostSelfCollision}{U_\text{s}}

\newcommand{\Spheres}{\boldsymbol{S}} 

\newcommand{\xsphere}{x} 
\newcommand{\rsphere}{r}

\newcommand{\net}{\Theta}
\newcommand{\dataset}{G}
\newcommand{\datain}{x}
\newcommand{\dataout}{y}

\newcommand{\omp}{\mathrm{OMP}}

\newcommand{\pperc}{p_{\mathrm{perc}}}
\newcommand{\pratio}{p_{\mathrm{ratio}}}
\newcommand{\ulist}{V}

\newcommand{\FB}{B}
\newcommand{\fb}{b}

\newcommand{\timewp}{t}
\newcommand{\timesub}{u}

\newcommand{\Ntime}{N_{\mathrm{\timewp}}}
\newcommand{\Ntimesub}{N_{\mathrm{\timesub}}} 
\newcommand{\Njoints}{N_{\mathrm{\qconfig}}} 
\newcommand{\Nbasisset}{N_{\mathrm{b}}} 
\newcommand{\Nframes}{N_{\mathrm{f}}} 
 
\newcommand{\Nspheresi}{N_{\mathrm{s}i}} 
\newcommand{\Nspheresj}{N_{\mathrm{s}j}} 
\newcommand{\Ndataset}{N_{\mathrm{\dataset}}} 

\newcommand{\WorldOccupancy}{W_{\mathrm{O}}} 
\newcommand{\WorldPointCloud}{W_{\mathrm{P}}} 
\newcommand{\WorldBasisSet}{W_{\mathrm{B}}}

\definecolor{red_flow}{HTML}{CC0000} 
\definecolor{green_flow}{HTML}{009900} 
\definecolor{blue_flow}{HTML}{0066CC}

\usepackage{fancyhdr}
\fancypagestyle{FirstPage}{
\lfoot{\fontsize{7}{7}\selectfont \begin{center}\copyright2022 IEEE. Personal use of this material is permitted. \\
Permission from IEEE must be obtained for all other uses, in any current or future media, including reprinting/republishing this material for advertising or promotional purposes, creating new collective works, for resale or redistribution to servers or lists, or reuse of any copyrighted component of this work in other works.\end{center}}
}

\title{\LARGE \bf
Speeding Up Optimization-based Motion Planning\\
through Deep Learning
}

\author{Johannes Tenhumberg$^{1,2}$, Darius Burschka$^{3}$ and Berthold Bäuml$^{1,2}$
\thanks{$^{1}$DLR Institute of Robotics \& Mechatronics, Germany;
$^{2}$Deggendorf Institute of Technology, Germany; 
$^{3}$Technical University of Munich, Germany}
\thanks{Contact: \tt\footnotesize johannes.tenhumberg@dlr.de}
\thanks{Supported by the Bavarian Ministry of Economics, project SMiLE2gether.}
}

\begin{document}
\maketitle
\thispagestyle{empty}
\pagestyle{empty}


\begin{abstract}
Planning collision-free motions for robots with many degrees of freedom is challenging in environments with complex obstacle geometries. 
Recent work introduced the idea of speeding up the planning by encoding prior experience of successful motion plans in a neural network.
However, this ``neural motion planning" did not scale to complex robots in unseen 3D environments as needed for real-world applications.
Here, we introduce ``basis point set", well-known in computer vision, to neural motion planning as a modern compact environment encoding enabling efficient supervised training networks that generalize well over diverse 3D worlds.
Combined with a new elaborate training scheme, we reach a planning success rate of 100\,\%.
We use the network to predict an educated initial guess for an optimization-based planner (OMP), which quickly converges to a feasible solution, massively outperforming random multi-starts when tested on previously unseen environments. 
For the DLR humanoid Agile Justin with 19\,DoF and in challenging obstacle environments, optimal paths can be generated in 200\,ms using only a single CPU core. 
We also show a first successful real-world experiment based on a high-resolution world model from an integrated 3D sensor.
\end{abstract}

\section{Introduction}\label{sec:Introduction}
\thispagestyle{FirstPage}
At the core, robotic motion planning is about getting from the start joint configuration to a goal configuration while avoiding obstacles in the environment and self-collision along the path.
Solving a motion task fast and efficiently does not only mean that the robot spends less time contemplating and more time moving.
If a solver can find a feasible solution in a fraction of a second, there opens up the door for more reactive planning and integrating the global planner more tightly into the sensor/vision-action loop.
Furthermore, we can tackle more high-level tasks with multiple smaller motion problems if each substep can be solved efficiently.

An interesting approach for speeding up motion planning is not to solve each planning problem anew but to use experience from having solved related motion tasks before. 
Important for the applicability of such experience-based planners to real-world problems is that they not only allow for arbitrary start and goal configurations but also for arbitrary environment geometries as an input. 
So, ``related tasks" only means that the robot geometry is the same. 

\begin{figure}[t]
    \centering
	\includegraphics[width=0.80\linewidth]{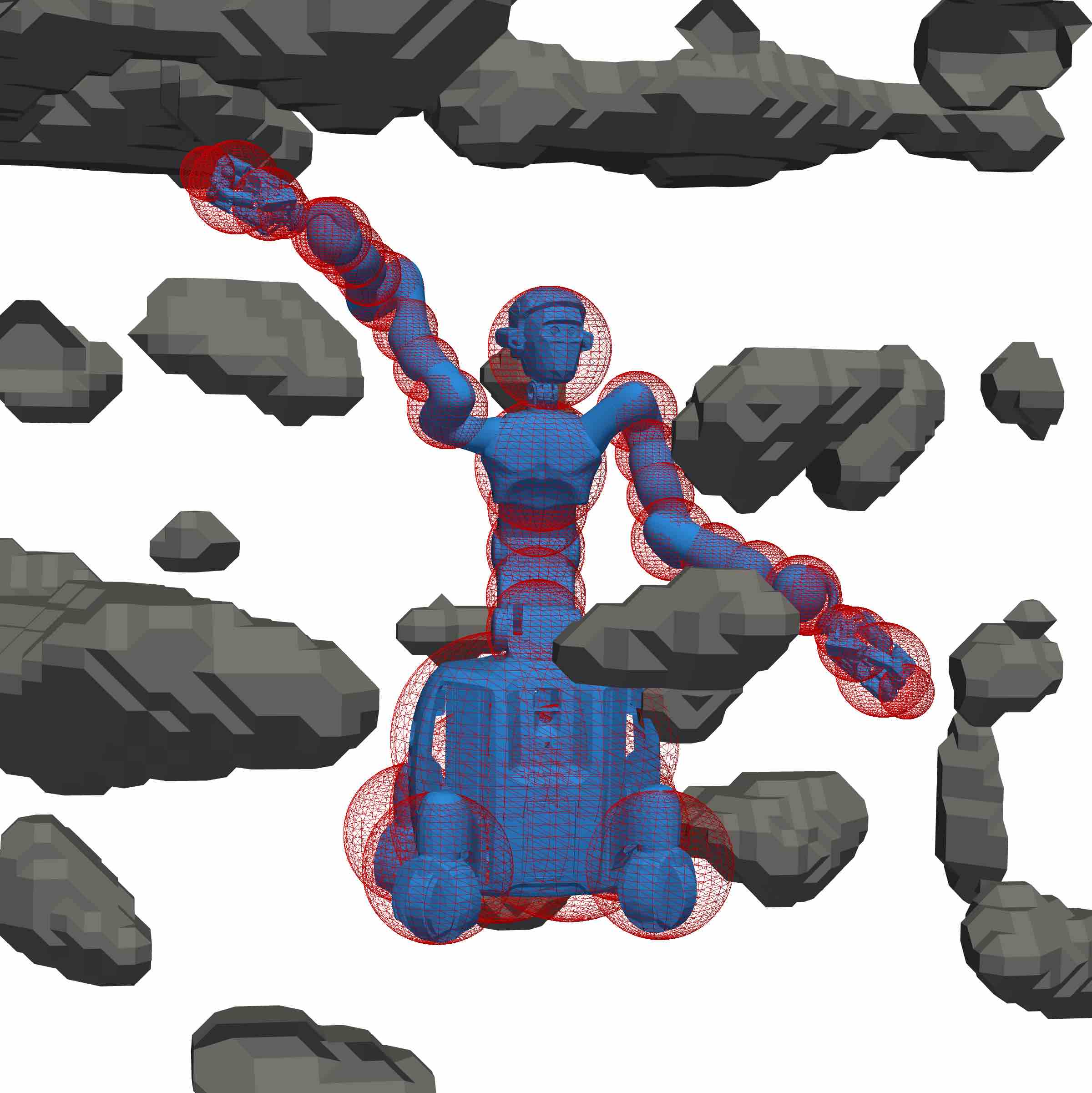}
	\vspace*{-3mm}
	\caption{
DLR's Agile Justin~\cite{Bauml2014} in a challenging obstacle environment, we generate with simplex noise for training and testing.
The sphere model of the 19\,DoF humanoid is shown as red wireframes. 
For videos of the motions, visit \href{\gitlink}{\gitlink}.}
	\label{fig:Justin_in_SimplexNoise}
	\vspace*{-5mm}
\end{figure}

This paper presents a deep learning enhancement for an optimization-based planner that allows robot motion planning in high-resolution environments with complex obstacle geometries.
For the DLR's humanoid robot Agile Justin~\cite{Bauml2014} with 19 DoF, feasible trajectories can be computed in only 200\,ms on a single CPU core (see \cref{fig:Justin_in_SimplexNoise}).

\subsection{Related Work}\label{sec:RelatedWork}

Two popular but fundamentally different approaches to solving a motion task in robotics are sampling-based planners (SMP) and optimization-based motion planners (OMP). 
SMP~\cite{LaValle2006} use randomness at their core and can guarantee to globally find a feasible, i.e., collision-free, path (if there is any). 
Examples are rapidly exploring random trees (RRT) and their extensions, which explore the configuration space iteratively and build a graph of possible configurations until a branch finds the goal. 
As the search space grows exponentially with the robots' degrees of freedom (DoF), vanilla variants of SMP don't scale well to complex robots.

In OMP, the motion task is formulated as an optimization problem~\cite{Zucker2013, Kalakrishnan2011, Schulman2014} and gradient-based techniques are used for non-convex optimization to find a solution. 
The computational complexity of those methods can, in principle, scale linearly with the DoF\footnote{Think of gradient descent where for each iteration only the computation of the gradient and its addition to the current joint angels are performed  -- both operations linear in the DoF when using, e.g., automatic differentiation.} and converge quickly to a smooth trajectory.
However, they only find a local minimum and strongly depend on the initial guess.

For non-trivial robot kinematics and environments, the objective function usually has many local minima.
Therefore, a multi-start approach is needed to find a feasible global solution, massively slowing down OMP in complex scenarios. 
STOMP~\cite{Kalakrishnan2011} and CHOMP~\cite{Zucker2013} try to mitigate the strong dependency on an initial guess by introducing stochasticity to the optimization.
TrajOpt~\cite{Schulman2014} uses convex hulls to represent the robot and its environment and increases the attraction basin for each optimum. 
But still, OMP needs multi-starts to find a feasible global solution.

OMP can be sped up by using experience from previously solved motions tasks by providing an \emph{educated} initial guess.
\citet{Jetchev2013} first proposed to save a database of feasible trajectories and look up a reasonable first guess for a new problem. 
\citet{Merkt2018} improved the idea through more efficient database storage and tested it on a humanoid robot.
But for both methods, only results for fixed or only slightly varying environments are shown.

Instead of databases, neural networks are often used to encode the experience.
The expectation is that they are more memory efficient, encode various solutions implicitly,  learn a general understanding of feasible trajectories, and produce useful predictions in unseen settings.
\citet{Qureshi2019, Qureshi2021} coined the term Motion Planning Network (MPNet) and then improved on the idea.
They used an RRT planner to collect feasible trajectories in 2D and 3D worlds and encode the environment with point clouds.
Their method works on the Baxter robot over ten known table scenes with 1000 paths per scene.
\citet{Strudel2020} showed that they could outperform these results by employing the PointNet~\cite{Qi2017} architecture to encode the point clouds of the environments. 
They achieved good results in 3D with a sphere and a rigid S-shape with three translational and rotational DoFs but did not consider a robotic application.
\citet{Bency2019} and \citet{Lembono2020} applied variations of the idea successfully to the two humanoids, Baxter and PR2.
However, they only used a single fixed environment without generalization to different worlds.
\citet{Lehner2018} trained a Gaussian mixture model to steer the search in a probabilistic roadmap.
The approach was demonstrated on a real 7 DoF robot but equally only for one fixed world.

Also, other learning methods have been applied to this problem.
For example, \citet{Jurgenson2019} used reinforcement learning with convolution layers to process the occupancy map of the world for 2D serial robots.
They invoke a classical planner only for cases where random exploration fails to find a feasible solution, so they implicitly use this expert knowledge to guide the training. 
\citet{Pandy2020} introduces a different approach, where the dataset generation is skipped entirely, and the network is directly trained using the objective function of the planning problem as the training loss, i.e., no supervision is used.
However, they only use geometric primitives to represent environments with few obstacles, limiting the flexibility.

For a more detailed overview of the literature, also refer to \citet{Surovik2020}. 
They introduce the term ``data-driven trajectory initialization" (DDTI) to generalize the ideas of ``trajectory prediction" or ``memory of motion". 
To our best knowledge, up to now, there is no experienced-based method for speeding up motion planning that can predict feasible paths for a complex robot in a large set of challenging and previously unseen 3D words.

\subsection{Contributions}\label{sec:Contributions}
\begin{figure}[tb]
    \centering
	\includegraphics[width=\linewidth]{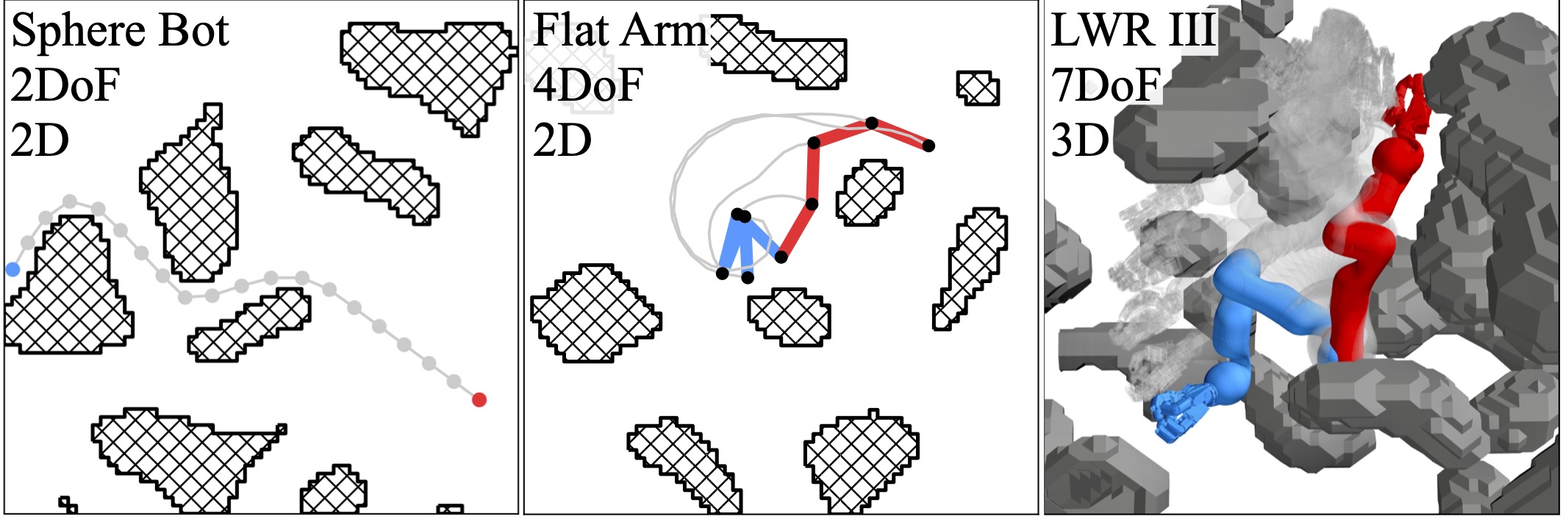}
	\vspace*{-4mm}
	\caption{
Three different robots in random 2D and 3D environments, generated with simplex noise.
A motion problem is described by the world (hatched),
the start configuration (blue), and the goal configuration (red).
The solution, the shortest feasible path from start to goal, is drawn in gray.
See \cref{tab:datasets} for an overview of the different robots and worlds in numbers.}
	\label{fig:3robots}
\end{figure}

We propose a neural network that encodes the experience of optimal trajectories over various tasks and worlds. 
We train the network supervised to predict the correct path for a given motion problem consisting of a world (obstacle environment) and start and end configurations. 
We then use the prediction of the network to warm-start an OMP, which ensures feasibility and smoothes the path.
Our main contributions are:

\begin{itemize}
\item Introducing substeps into CHOMP\cite{Zucker2013} to explicitly calculate the swept volume to facilitate that no collisions are missed between discrete waypoints.
\item The introduction of the basis point sets (BPS), described by \citet{Prokudin2019}, into learning-based motion planning.
This memory- and computational-efficient encoding enables training and generalization for complex robots in challenging environments.
\item Building new demanding datasets of motion tasks and optimal solution paths (via OMP) for training and testing. 
The dataset includes autogenerated random worlds with configurable complexity.
\item A training scheme, where we combine the network and the objective function as a metric to efficiently clean, extend and boost an initial dataset.
\item Extensive experiments in simulation for different robots ranging from a simple 2D sphere bot up to the 19 DoF humanoid Agile Justin in 3D. 
For Agile Justin, we also report a first real-world experiment, showing that Sim2Real transfer is working.
\end{itemize}

\section{CHOMP-like Motion Planning}\label{sec:OMP}

\begin{figure}[t]
    \centering
	\includegraphics[width=1.0\linewidth]{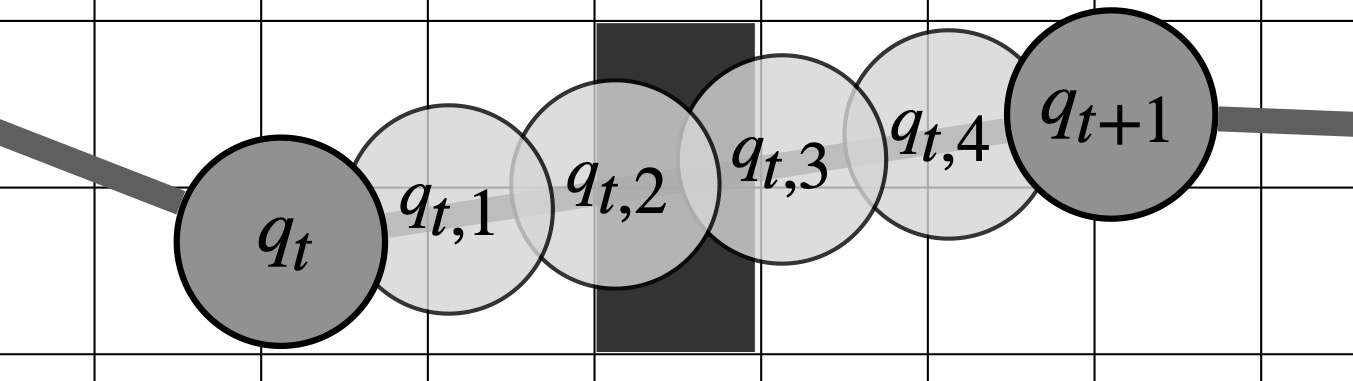}
	\caption{
Substeps $\qconfig_{\timewp, \timesub}$ between two discrete waypoints $\qconfig_{\timewp}$ and $\qconfig_{\timewp+1}$ to explicitly calculate the swept volume of the path in higher resolution.}
	\label{fig:subspheres}
\end{figure}

For generating the training samples for our network as well as for online post-processing, we use OMP. 
Ultimately, the neural network should do the primary workload in solving a motion task.
Therefore the runtime efficiency of the OMP is not of utmost importance. 
We implemented an OMP similar to CHOMP~\cite{Zucker2013}, where the robot is modeled as spheres and the environment by a signed distance field (SDF). 
A significant extension to CHOMP is that we introduce additional substeps interpolating between the time steps. 
This way, the swept volume can be computed to any arbitrary accuracy, i.e., guaranteeing to miss no obstacle, without increasing the number of optimization variables (see \cref{fig:subspheres}). 

OMP formulates the motion task of getting from point A to point B as an optimization problem with the desired path as the optimum. 
The path $\QPath$ consists of a discrete set of waypoints of joint configurations
\begin{align}
\QPath = [\qconfig_1, \dots, \qconfig_{\Ntime}], \, \qconfig_{\timewp} \in \mathbb{R}^{\Njoints},
\end{align}
and the task is encoded by an objective function  $\OMPCost(\QPath)$  measuring the quality of a given path.
While, in general, the specific objective can be chosen freely, the two usually used terms to get a collision-free and short path are
\begin{align}
\OMPCost(\QPath) = \OMPCostCollision(\QPath) + \lambda \OMPCostLength(\QPath).
\end{align}

The length cost $\OMPCostLength$ is given by 
\begin{align}
\OMPCostLength(\QPath) = \frac{\Ntime-1}{|\qconfig_{\Ntime}-\qconfig_1|^2} \sum_{\timewp=1}^{\Ntime-1} |\qconfig_{\timewp+1} - \qconfig_{\timewp}|^2,
\end{align}
which favors short and smooth trajectories. 
It is often convenient to scale the length cost by the minimal possible path length, the direct connection $|\qconfig_{\Ntime}-\qconfig_1|^2$.
Then, the shortest possible path always has a cost of one.

To calculate the collision cost between the robot and the environment, we need a model for both.
The forward kinematics $\Frames = \funkin(\qconfig)$ maps from joint configurations to the link frames $\Frames_i$ and each link's geometry  is described by a set of spheres $\Spheres_i=\{\xsphere_{ik}, \rsphere_{ik}\}_{k=1}^{\Nspheresi}$ with centers and radii.
The world is represented by a SDF $\distfield(x)$ which gives the distance to the closest obstacle for each point $x$ in the workspace.
The collision cost is then given by the sum of all the collisions of the different body parts along the path
\begin{align}
\OMPCostCollision(\QPath) &= \sum_{\timewp, \timesub}^{\Ntime, \Ntimesub} \sum_{i=1}^{\Nframes} \sum_{k=1}^{\Nspheresi} 
\funclip\Big( 
    \distfield\big( 
        \Frames_i(\qconfig_{\timewp, \timesub}) \cdot{} \xsphere_{ik}
    \big) 
    - \rsphere_{ik}
\Big),\\
\qconfig_{\timewp, \timesub} &= \qconfig_{\timewp} + \frac{u}{\Ntimesub}(\qconfig_{\timewp + 1} - \qconfig_{\timewp}).
\end{align}
In extension to the original CHOMP algorithm~\cite{Zucker2013}, we subdivide the path between two waypoints into substeps $\qconfig_{\timewp, \timesub}$ via linear interpolation (see \cref{fig:subspheres}). 
This way, a collision-free path can be guaranteed, when the number of substeps $\Ntimesub$ is adjusted to the step length and the voxel size of the SDF. So, the swept volume of each sphere is \emph{explicitly} computed instead of the \emph{implict} computation CHOMP performs via a projection of the cartesian velocity vector of the moving spheres. 

The smooth clipping function only considers the parts which are in collision by setting positive distances to 0.
Thus, a collision-free path has a collision cost of 0.
\begin{align}
\funclip(d) &= 
\left\{
\begin{array}{ll}
-d + \frac{\epsilon}{2}               & \text{, if }                d <    0 \\ 
\frac{1}{2 \epsilon} (d - \epsilon)^2 & \text{, if }  0        \leq d \leq \epsilon \\
0                                     & \text{, if }  \epsilon <    d
\end{array}
\right.
\end{align}
In addition to collisions with the world,  complex robots must also account for self-collision.
Again, the cost sums up all the penetrations between the different body pairs 
\begin{equation}
\begin{aligned}
\OMPCostSelfCollision(\QPath) =& \sum\nolimits_{\timewp, \timesub}^{\Ntime, \Ntimesub} \, \sum\nolimits_{j>i}^{\Nframes, \Nframes} \, \sum\nolimits_{k, l}^{\Nspheresi, \Nspheresj} \\
&\funclip\big( 
    \big| 
         \Frames_{i}(\qconfig_{\timewp, \timesub}) \cdot \xsphere_{ik}  -  \Frames_{j}(\qconfig_{\timewp, \timesub}) \cdot \xsphere_{jl} 
    \big|
    - \rsphere_{ik} - \rsphere_{jl}
\big).
\end{aligned}
\end{equation}
Here again, we use substeps to guarantee collision-free paths via explicitly checking the swept volume. 
To our knowledge, no other OMP-based planner is doing this.

With this formulation, the optimal path and solution to the motion task is the one with the lowest objective
\begin{align}
\QPath^* = \argmin_{\QPath} \OMPCost(\QPath).
\end{align}
We use gradient descent for iteratively finding a minimum of the objective function, starting with the initial guess $\QPath_0$,
\begin{align}
\QPath_{i+1} = \QPath_i - \alpha \frac{\partial \OMPCost(\QPath_i)}{\partial \QPath_i}.
\end{align}
We use vanilla gradient descent as efficiency in the OMP part is not the primary concern of our method and it allows for easy adaption and parameterization (only $\alpha$)

\section{Dataset Adaption for Efficient Learning}\label{sec:Experience}

\begin{figure}[tb]
    \centering
	\includegraphics[width=1.0\linewidth]{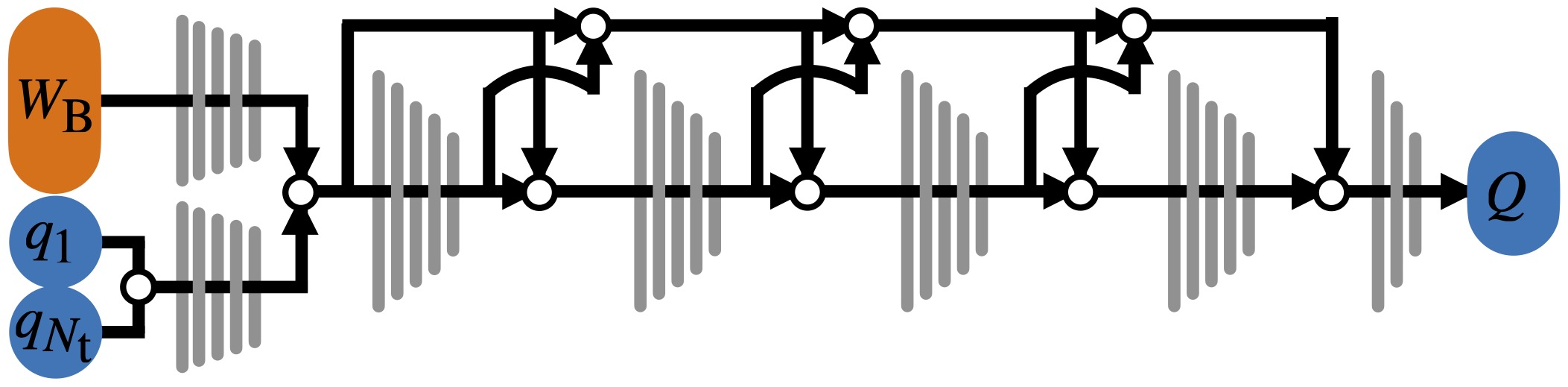}
	\vspace*{-4mm} 
	\caption{
Network architecture to map from a given motion task (world, start and end configuration) to an optimal path.
Blocks of tapered Fully Connected Layers (gray) are combined like the DenseNet architecture~\cite{Jegou2016} via skip-connections and concatenations.
See the bottom of \cref{tab:datasets} for the number of network parameters used for the different robots.}
    \label{fig:network}
\end{figure}

The idea is to no longer rely on random multi-starts and speed up the planning time by using a neural network to predict an initial guess for OMP. 
The network should encode the experience of successful paths by learning a mapping from a motion task, consisting of a world and a start $\qconfig_1$ and end configuration $\qconfig_{\Ntime}$, to the intermediate waypoints of an optimal path $\QPath^* = [\qconfig_2, ..., \qconfig_{\Ntime-1}]$.
\cref{fig:network} shows the network architecture we use.
Besides encoding the in- and output of the network, a crucial point for supervised learning is the dataset. 
The following section discusses several insights into the generation and usage of such a dataset with and for OMP.
Our methods substantially increase the final prediction quality of our network and make training more efficient. 
See \cref{sec:ResultsDataset} for experimental validation of these methods.

\subsection{Challenging Samples}
The training data distribution should represent the actual application and focus on challenging examples.
Suppose the dataset is too easy, and direct linear connections from A to B drastically outweigh more complex trajectories. 
In that case, training the network can quickly get stuck in a local minimum and predict only straight lines, regardless of the given task. 
One possibility of generating a dataset with more challenging motion problems is to consider only samples where OMP using the direct connection as an initial guess does not converge to a feasible path.

\subsection{Consistent Samples}
Besides finding challenging samples, the ambiguity of motion planning (more than one feasible solution) can become a problem for trajectory regression.
Even if we assume that we can resolve the ambiguity of the optimal path via the objective for the shortness of the path, there will be ``close calls" in the dataset (almost the same objective values but fundamentally different paths).
Furthermore, the classical planner using a limited number of multi-starts can only produce suboptimal labels, making it hard for the network to learn consistent mapping.
This is especially true for challenging tasks where the classical planner often fails and only produces feasible paths in a small fraction of the tries.

\subsection{Symmetries of Motion Planning}
\begin{figure}[t]
    \centering
    \includegraphics[width=1.0\linewidth]{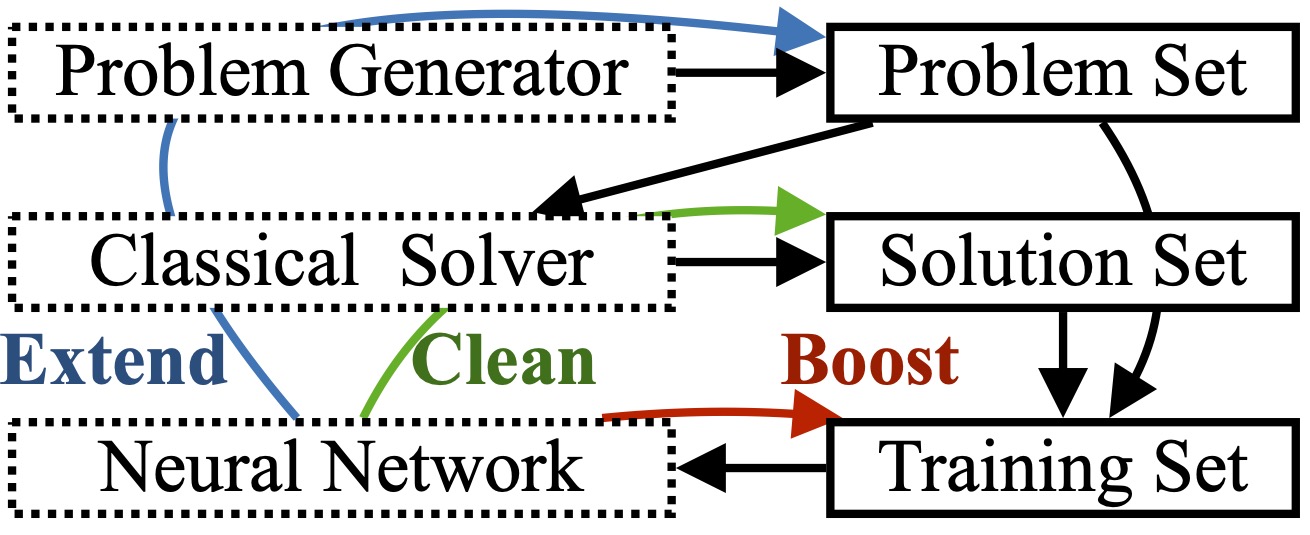}
	\caption{
Scheme showing the connection between non-learning-based solver, dataset, and neural network.
The colored arrows indicate the information flow for cleaning, extending, and boosting the dataset with the guidance of the network.
Cleaning: use the solver to update labels in the solutions set; 
Boost: overrepresent existing hard examples in training; 
Extend: generate new hard examples for the network and add them to the problem set.
}
	\label{fig:flowchart}
\end{figure}

Data generation is costly, so one can use symmetries in motion planning to efficiently use the information in each sample.
If one has the optimal path from A to B, one also has the solution from B to A. 
This assumption is no longer valid if terms in objective function break the temporal symmetry.
Furthermore, many robots also have spatial symmetry axes.
Often, it is possible to align the first joint of the robot with one axis of the cartesian coordinate system of the environment.
Doing so allows us to rotate the world and this first joint simultaneously without changing the optimality of the resulting trajectory in the new world.
\citet{Chamzas2019} use this spatial symmetry to store paths in their database more efficiently.
While ideally, one wants to integrate these symmetries directly in the data representations or the network architecture, we used them to augment and increase the dataset.

\subsection{Interplay between Network and Dataset}

\begin{algorithm}[t]
    \begin{algorithmic}

\Procedure{main}{}
    \State create initial dataset $\dataset = \{(\datain_i, \dataout_i)\}^{\Ndataset}_{i=1}$ with OMP
    \State train net $\net$ on dataset $\dataset$ 
    \While{improvement on testset}
        \State CleanDataset($\dataset, \net$)
        \State ExtendDataset($\dataset, \net$, $N=| \dataset |/20$)
        \State BoostDataset($\dataset, \net$, $\pperc=0.8$, $\pratio=0.9$)
        \State train net $\net$ on dataset $\dataset$
    \EndWhile
\EndProcedure
\Procedure{CleanDataset}{$\dataset$, $\net$} \Comment{{\color{green_flow}Clean}}
    \For{$(x_i, y_i)$ in $\dataset$} 
        \State $y_p \gets \net(x)$
        \State $y_p^* \gets \omp(x, y_p)$
        \If{$\OMPCost(y_p^*) \leq \OMPCost(y)$}
            \State $\dataset.\mathrm{replace}(y \gets y_p^*$)
        \EndIf
    \EndFor
\EndProcedure
\Procedure{ExtendDataset}{$\dataset$, $\net$, $N$} \Comment{{\color{blue_flow}Extend}}
    \For{$k \gets 1$ to $N$}
         \State $x_i \gets \mathrm{sampleNewProblem()}$
         \If{$\ulist_p  < \OMPCost(\net(x_i))$}
             \State $y_i \gets \omp(x_i)$
             \State $\dataset.\mathrm{append}((x_i, y_i))$
         \EndIf
    \EndFor
\EndProcedure
\Procedure{BoostDataset}{$\dataset$, $\net$, $\pperc$, $\pratio$} \Comment{{\color{red_flow}Boost}}
    \State $\ulist \gets [\OMPCost(\net(x_i))$ for $(x_i, y_i)$ in $\dataset$]
    \State $\ulist_p \gets$ percentile($W, \pperc$)
    \For{$(x_i, y_i)$ in $\dataset$}
        \If{($\OMPCost(\net(x)) < \ulist_p$ and random(0, 1) $< \pratio$)}
            \State $\dataset.\mathrm{remove}((x_i, y_i))$
        \EndIf
    \EndFor
\EndProcedure
                    
\end{algorithmic}

    \caption{Improvement of network and dateset}
    \label{alg:network+dataset}
\end{algorithm}

We propose to use the neural network $\net$ to correct and enhance its own training data $\dataset = \{(\datain, \dataout)\}$.
This approach is possible whenever synthetic data is used for training, and one has an objective metric to measure the quality of a prediction.
The assumption is that the network can learn some aspects of the problem even on an imperfect dataset and its predictions become better than random guessing.  

When generating a dataset with and for OMP, we can use the objective $\OMPCost(\QPath)$ as a universal quality metric.
The idea of interweaving the network training closer with the dataset generation and improvement is summarized in \cref{fig:flowchart} and described by \cref{alg:network+dataset}. In what follows, we give a detailed explanation of the different methods.

\subsubsection{Clean}
First, the network can be used as guidance to double-check where the dataset is inconsistent. 
If the label and the network's prediction $\QPathNet$ are close, and the respective objective $\OMPCost(\QPathNet)$ is small, no action is necessary.
However, if there is a discrepancy between prediction and label, it is worthwhile to use more multi-starts with the OMP to get a better label for the sample. 
If a prediction has a better objective than the current label, we can replace it without adding any bias to the dataset.
Doing so will improve the labels and make the dataset more consistent.
This makes it easier for the network to find the underlying patterns.

\subsubsection{Boost}
After some training, the network has learned to predict good paths for the relatively simple samples, but the more challenging outliers are still not solved.
Therefore, we use boosting to select challenging samples with higher probability during training, increasing the incentive to learn these samples. 
We steer this kind of curriculum again by using $\OMPCost$ as a metric: the higher the difference between the predicted and actual cost for a given sample is, the more challenging it is.
In our experiments, the challenging tasks for the network correlated well with the relative path length and the number of obstacles in the scene.

\subsubsection{Extend}
Lastly, one can use the network to generate new samples.
The idea is to use the network's performance on a new sample as a metric for information gain.
To improve the network, one wants specifically to add samples where the network performs poorly.
To decide this before spending resources to produce a new label using OMP, one can use the objective of the prediction $\OMPCost(\QPathNet)$.
If it is small, the network can solve this task already.
However, if the objective is significant, the task is challenging for the network, and we include it in the training set.

\section{Basis Point Set as Efficient World Encoding}\label{sec:BasisPointSet}

\begin{figure}[tb]
    \centering
	\includegraphics[width=\linewidth]{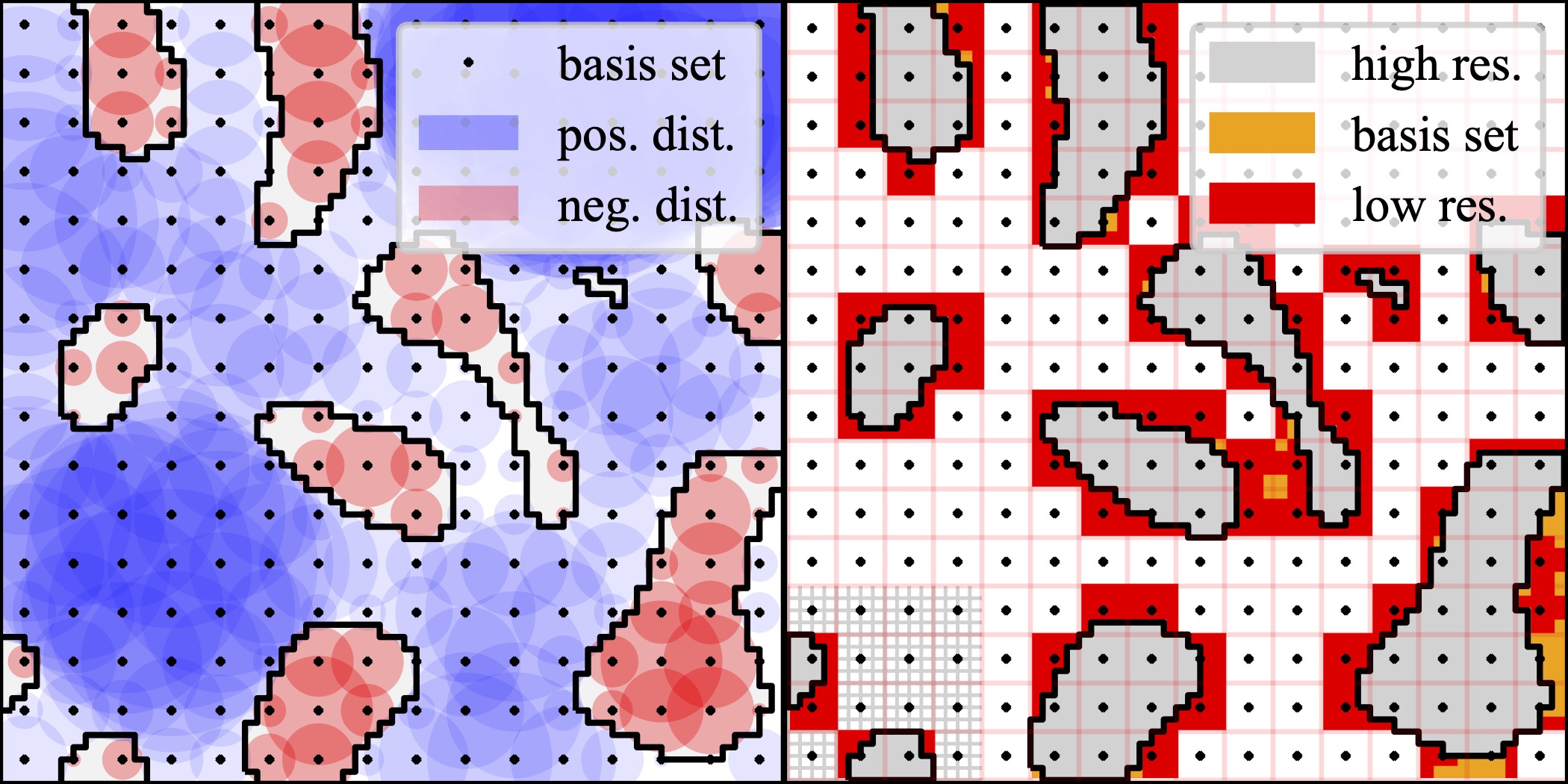}
	\vspace*{-3mm}
	\caption{
Visualization of the representation capabilities of basis point sets. 
On the left, a regular BPS with $16\times16=256$ points and their respective distances to the closest obstacle in an occupancy map with $64\times64 =4096$ pixels is shown.
Blue areas describe positive distances to obstacles and are guaranteed free of obstructions.
Red regions show negative distances and are completely inside barriers.
We can draw no direct conclusion from the white areas, but they have to be marked as obstacles to be conservative. 
The BPS representation for this regular grid is equivalent to subsampling a high resolution $64\times64$ SDF.
The right image shows the reconstruction of the BPS to the full image with the errors marked in orange. 
A patch of the original grid is added to highlight the difference in resolution.
The far more conservative result of reducing the resolution of the occupancy grid by the same factor is highlighted in red.
This comparison demonstrates that the basis set preserves more information than conservatively shrinking the occupancy grid by the same factor. 
A further advantage of the untruncated distances is that even if there are no nearby data points close for a basis point, it can nevertheless help in representing a surface further away.
}
    \label{fig:fb_exact}
\end{figure}

If a network should speed up the OMP approach with a valuable warm-start, it needs to ``understand" the robot motion in arbitrary unseen environments.
Hence, a suitable encoding of the world is essential.

Two central environment representations in robotics and computer vision are occupancy grids $\WorldOccupancy$ and point clouds $\WorldPointCloud$.
Both have their advantages and disadvantages. 
While occupancy grids can be processed like images in 2D, their memory inefficiency and the high computational cost for the convolution operations become a burden in 3D. 
On the other hand, point clouds are a denser representation. 
Still, they have no fixed size and no inherent ordering, making it hard for a network to learn a permutation invariant mapping.

\citet{Prokudin2019} recently introduced a new idea to represent spatial information in computer vision, which is especially suited for deep learning. 
Choosing a fixed set of basis points once and measuring the distances relative to this set for all new environments does not have the problem of varying permutations and lengths as point clouds have. 
It is also far more efficient in terms of memory and computation than voxel grids, allowing fast training in high-resolution 3D environments.

Formally, the BPS is an arbitrary but fixed set of points
\begin{align}
\FB = [\fb_1, \dots, \fb_{\Nbasisset}], \,\, \fb_i \in \mathbb{R}^d.
\end{align}
The feature vector $\WorldBasisSet$ passed to the network consists of the distances to the closest point in the environment for all basis points.
If the environment is defined by a point cloud $\WorldPointCloud$ this can be calculated by 
\begin{align}
\WorldBasisSet = [\min_{x_i \in \WorldPointCloud} |b_1 - x_i|, \dots ,  \min_{x_i \in \WorldPointCloud} |b_{\Nbasisset} - x_i|].
\end{align}
Alternatively, if the environment is given by an occupancy grid or a distance field $\distfield$ like we used for OMP in \cref{sec:OMP}, one can directly look up the feature vector
\begin{align}
\WorldBasisSet = [\distfield(\fb_i), \dots , \distfield(\fb_{\Nbasisset})].
\end{align}
With the second approach, it is possible to use signed distances. 
This adds directly a notion of inside and outside to the representation of the world. 

\begin{figure}[tb]
    \centering
	\includegraphics[width=\linewidth]{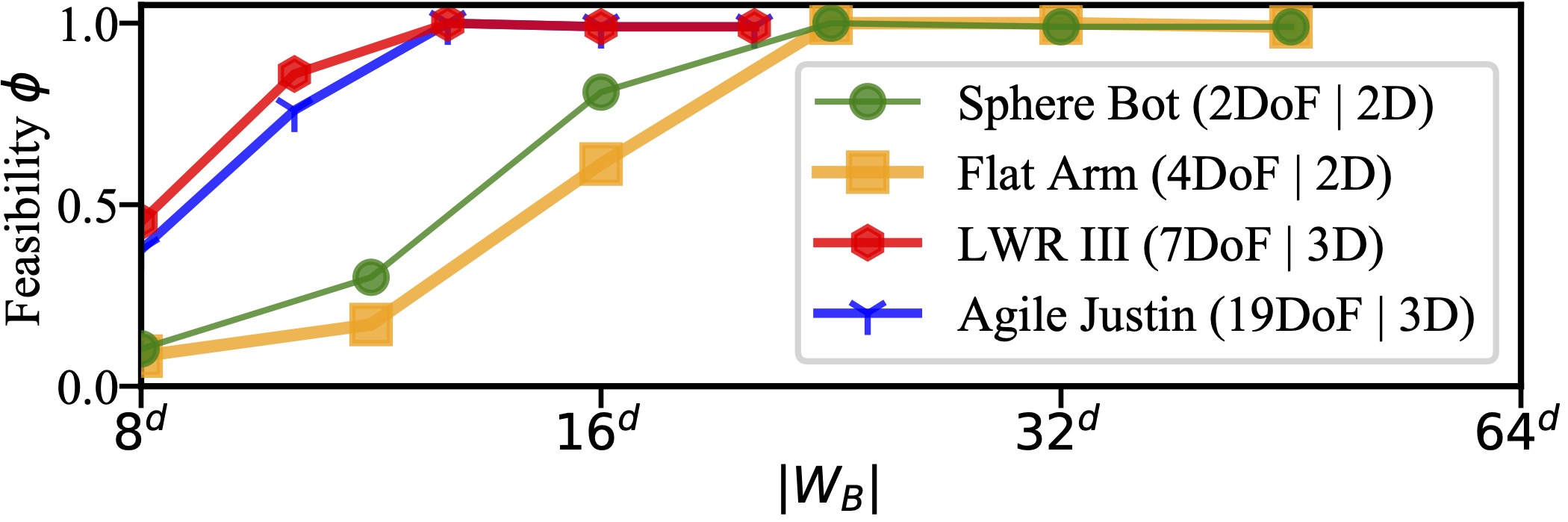}
    \vspace*{-5mm}
	\caption{
Influence of the size of BPS $|\WorldBasisSet|$ on the feasibility $\Feasibility$ of the network prediction after OMP.  
The $|\WorldBasisSet|$ on the x-axis is scaled by the world dimension.
Too few points can't represent the obstacles in enough detail for the network to make useful predictions.
However, the required number of points is significantly smaller than the resolution of the underlying occupancy grid ($64^d$).
It was not feasible to increase the input layer to $64^3$ basis points in 3D.
}
	\label{fig:number_of_bps}
\end{figure}

Before using this representation to train a neural network, in \cref{fig:fb_exact} we analyze its properties with regard to efficiency and safety in a setting without learning.   

Although an occupancy grid directly represents the cartesian workspace, the mapping to a robot's configuration space can be complex. 
But the planning network has to understand this mapping to connect movements in joint space with its implications in the  world to be able to plan collision-free paths.
In \cref{sec:Results}, we show in experiments that the basis point set enables to learn that connection for complex robot kinematics and diverse environments.

\section{Experimental Setup}\label{sec:Experiments}

\begin{figure}[t]
    \centering
	\includegraphics[width=\linewidth]{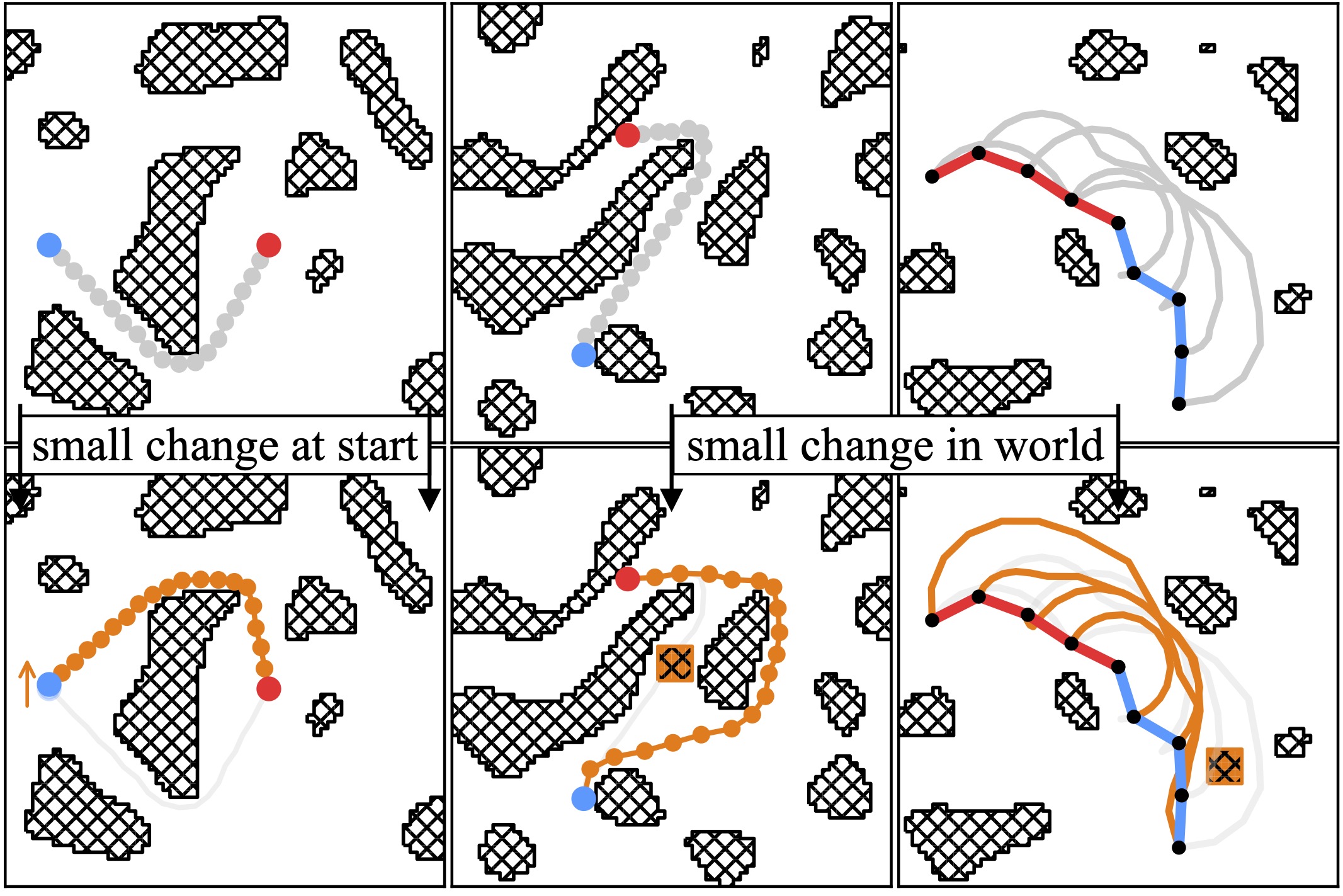}
	\vspace*{-3mm}
	\caption{
Three examples of the networks' prediction for only slightly altered problems.
The changes in the start configuration (left) and the environment (middle, right) are highlighted in orange. 
The networks' predictions after optimization for those new problems are shown in the bottom row, indicating they can react sharply to small input changes.
For easier visibility in 3D, we refer to animations for the other robots on the website}
\label{fig:NetworkBehaviour}
\end{figure}

\subsection{Dataset}

We constructed extensive datasets for different robots (see \cref{fig:Justin_in_SimplexNoise,,fig:3robots,,fig:NetworkBehaviour}) to validate our methods.
For 2D, we investigated  a simple sphere robot with 2\,DoF and a serial arm with 4\,DoF.
Furthermore, we used two real 3D robots:
the LWR III~\cite{Hirzinger2002}, a robotic arm with 7\,DoF, and humanoid robot DLR Agile Justin~\cite{Bauml2014} with 19\,DoF distributed over an upper body and two arms.

To generate diverse and challenging worlds, we used simplex noise~\cite{Gustavson2005}.
This gradient noise is used in video games to create random but naturally-looking levels. 
A typical example is a continuous height map.
By changing the cut-off threshold and the resolution of this noise, we can vary the density and form of the obstacles in the binary occupancy grid.
To ensure that the environments are not too densely packed with obstacles, at least 200/1000 random robot configurations $\qconfig$ must be feasible to include a world in the dataset.
Examples can be seen in \cref{fig:Justin_in_SimplexNoise,,fig:3robots}.

To generate the labels for the samples, we used the OMP approach described in \cref{sec:OMP}, with naive gradient descent and fixed step size.
The paths consist of $\Ntime=20$ waypoints, and to make the dataset challenging, we only included hard tasks that were not solvable starting from a straight line as initial guess.
All other paths were discarded as too easy.
We used up to 100 multi-starts and always picked the shortest feasible solutions as the correct label. 
The heuristic for generating the initial guesses was to use 1 to 3 random points in the configuration space and connect them linearly with the start and endpoint.
See \cref{tab:datasets} for an overview of the robots and the datasets\footnote{
All datasets plus additional information on their generation and use in training are provided at \href{\gitlink}{\gitlink}.}.
The table also shows initially mislabeled paths and paths specifically included to challenge the network.

\subsection{Network}

The last lines of \cref{tab:datasets} show the network details in numbers and \cref{fig:network} displays the general architecture.
All the networks were trained purely supervised with a mean squared error between the predicted path $\QPathNet$ and the label $\QPath$ as loss function. 
As the encoding for the path,  the deviation from the straight line is used. 
This representation implies that even an untrained network producing only random noise around zero can make meaningful predictions.
For start and end, we use the normalized joint vectors $\qconfig_1$ and $\qconfig_{\Ntime}$ as input.
As environment encoding, we use the BPS described in \cref{sec:BasisPointSet}
with  a hexagonal closed packing and only consider points inside the robots' maximal reach.
See \cref{fig:number_of_bps} for an analysis of the dependency of the prediction quality on the size of the BPS.

\section{Evaluation Results}\label{sec:Results}

From the 10000 environments we generated for each robot, we used 9000 for the network's training and the remaining unseen worlds for testing. 
All the results in \cref{sec:Results} are based on this unseen test set with 10000 hard motion tasks.
As a quality measure we report the feasibility rate $\Feasibility$, i.e., the quotient of the number of feasible paths and the size of the test set.

\begin{table}[t]
    \caption{Overview of the datasets and nets for the different robots.}
    \resizebox{\linewidth}{!}{%
\begin{tabular}{c|cccc}
\toprule
                                      &  Sphere Bot                           &  Flat Arm                      &  LWR III                                &  Agile Justin \\
\midrule
DoF                                   &  2                                    &  4                             &  7                                      &  19                               \\
World size                            &  $10\times10\mathrm{\mathrm{\,m}}^2$  &  $1.0\times1.0\mathrm{\,m}^2$  &  $1.2\times1.2\times1.2\mathrm{\,m}^3$  &  $3\times3\times3\mathrm{\,m}^3$  \\
Grid dimensions                       &  $64 \times 64 $                      &  $64 \times 64$                &  $64 \times 64 \times 64$              &   $64 \times 64 \times 64$         \\
\# Worlds                             &  $10^4$                               &  $10^4$                        &  $10^4$                                 &  $10^4$                           \\
\# Paths                              &  $0.6\times10^6$                      &  $6.5\times10^6$               &  $2.2\times10^6$                        &  $3.7\times10^6$                  \\
Avg. time p. core                     &  0.1\,s                               &  0.8\,s                        &  3.1\,s                                 &  8.4\,s                           \\
\midrule                      
\# Improvements                       &  $0.1\times10^6$                      &  $0.3\times10^6$               &  $0.2\times10^6$                        &  $0.3\times10^5$                  \\
\# Extensions                         &  $0.5\times10^5$                      &  $1.5\times10^5$               &  $3.0\times10^5$                        &  $5.0\times10^5$                  \\
\midrule                      
Avg. $\OMPCostLength(\QPath)$         &  1.589                                &  1.7136                        &  1.551                                  &  1.483                            \\
Avg. Feas. $\Feasibility$             &  67.3\,\%                             &  32.6\,\%                      &  54.5\,\%                               &  44.1\,\%                         \\
Avg. time p. core                     &  0.1\,s                               &  0.8\,s                        &  3.1\,s                                 &  8.4\,s                           \\
\midrule
\midrule
Net \\ 
\# In $\rightarrow$ \# Out            &  $516 \rightarrow 39$                 &  $520 \rightarrow 72$          &  $2062 \rightarrow 126$                 &  $2086 \rightarrow 342$           \\
$|\WorldBasisSet|$                    &  512                                  &  512                           &  2048                                   &  2048                             \\
$N_\mathrm{t}$                        &  18                                   &  18                            &  18                                     &  18                               \\
\# Parameters               	      &  $3.4\times10^6$                      &  $7.5\times10^6$               &  $2.4\times10^7$                        &  $4.1\times10^7$                  \\
\bottomrule
\end{tabular}
}

    \label{tab:datasets}
\end{table}

\subsection{Dataset Adaption}\label{sec:ResultsDataset}

\cref{tab:ablation_study} shows the influence of the methods for dataset adaption during training as discussed in \cref{sec:Experience}.
As metric we use the feasibility rate $\Feasibility$ of the predicted paths after further iterations with the OMP as described in \cref{sec:OMP}.
First, we compare the hard and the easy dataset. 
Because the easy dataset consists only of paths produced from straight lines, the overall variance is too slight, and the network does not learn to avoid the obstacles.
This network is not able to solve the test set of hard examples.
Next, we introduced the different modes of data augmentation to increase the size of the dataset.
The temporal and spatial symmetries improve the feasibility rate $\Feasibility$ without additional computing costs.
The number of cleanings describes how often the labels were updated with the help of the neural network.
Each iteration brings the labels closer to the optimal solution and makes the dataset more consistent, leading to better results. 
At this stage, the network performs already well with a success rate of over 85\,\%, but there are still tasks the net can not solve. 
We add the boosting technique to overrepresent more challenging samples during training to increase the feasibility further.
As the final step, we use the trained network to generate more challenging samples.
With this approach, we achieved 100\,\% feasibility on the hard unseen test set.

\subsection{Comparison to Random Multi-start}

\begin{figure}[t]
    \centering
	\includegraphics[width=1.0\linewidth]{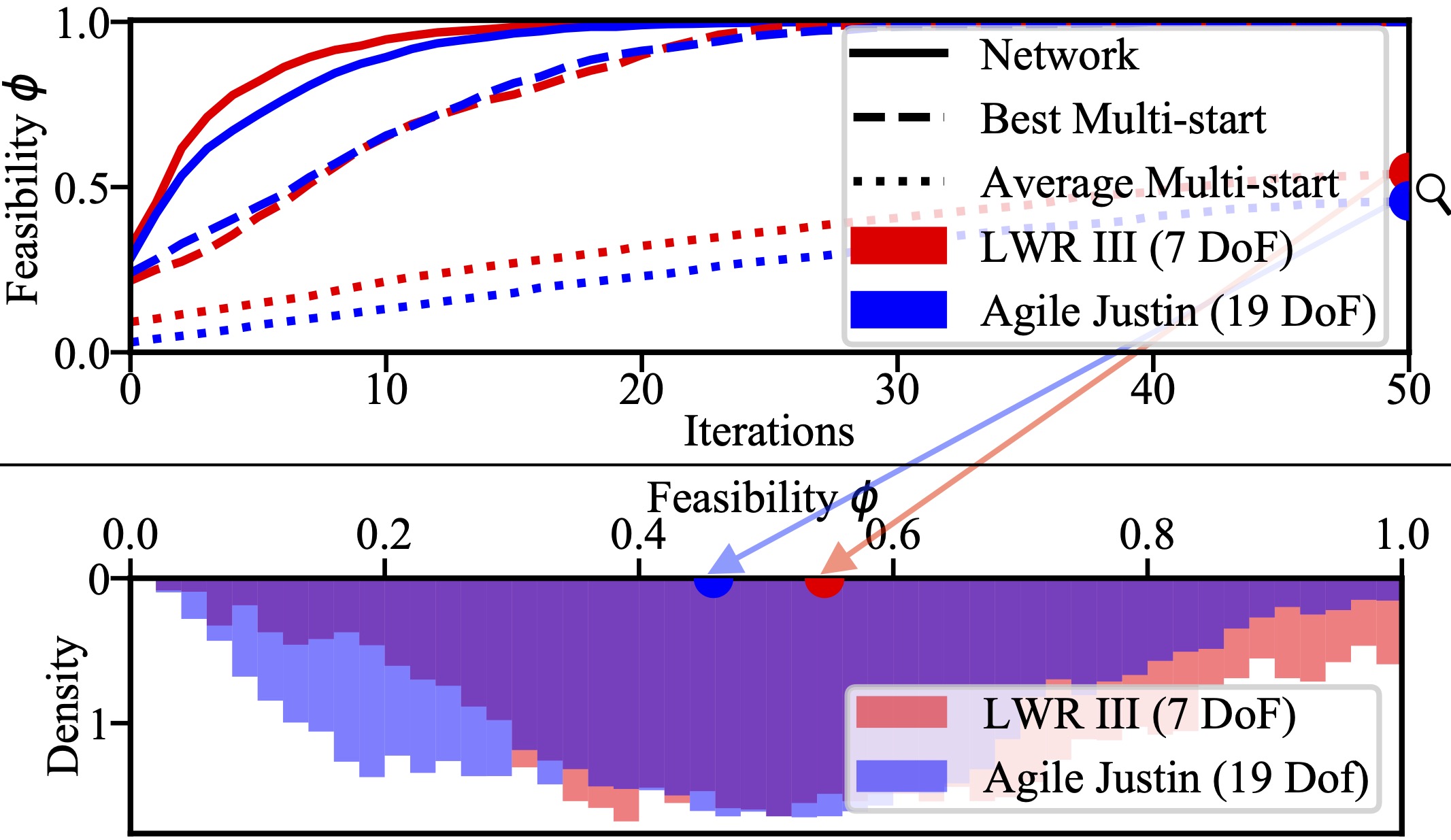}
	\vspace*{-4mm}
	\caption{Top: Average convergence to feasibility $\Feasibility$ of OMP for different initial guesses.
	The prediction of the network outperforms the average and even the best out of 100 multi-starts significantly.
	Bottom: Distribution of the average feasibility $\Feasibility$ of the random multi-starts after 50 OMP iterations.}
	\label{fig:results_iterations}
\end{figure}

The capabilities of our method become apparent when we compare the prediction of the network to the heuristic with random multi-starts used to create the dataset. 
In \cref{fig:results_iterations} the convergences to a feasible path of different initial guesses are displayed for the LWR III and Agile Justin.
Without any experience, the best one can do, is to try random multi-starts and hope one of them converges.
From the 100 multi-starts we used per task, only 50\,\% converge to a feasible solution after 50 iterations.
Even the lucky initial guess, which converged the fastest for each problem, gets outperformed by the network's prediction.
The crucial difference is that our network does not depend on chance but can reliable predict initial guesses that converge after a few iterations to a feasible solution. 

The actual speed gain is even more prominent when looking at the distribution over different motion tasks (see bottom of \cref{fig:results_iterations}).
There are problems for the LWR III and Agile Justin where only 10\,\% or less of all multi-starts converge to a feasible path.
If one wants to find a solution to such a problem with 90\,\% confidence, one would need more than $\log(1-0.1)/\log(1-0.9) > 20$ multi-starts, making the initial guess of the network effectively over 20 times faster.

On our test machine (Intel i9-9820X @ 3.30\,GHz, 32\,Gb RAM), a single iteration of gradient descent for one path of Agile Justin takes $10\,\mathrm{ms}$ on a single core.
Using the network's prediction as a warm-start and stopping each sample after convergence leads to an average run time of $182(\pm29)\,\mathrm{ms}$ with a worst-case of $334\,\mathrm{ms}$.

\subsection{World Encoding}

\begin{table}[t]
\caption{Analysis of different dataset distributions and extensions \& training methods for Agile Justin.}
\centering
\begin{tabular}{cc|cc|cc}
\toprule
\multicolumn{2}{c}{Dataset}  &  \multicolumn{2}{c}{Training}   &  \multicolumn{2}{c}{Feasibility $\Feasibility$} \\
\# Cleans  &  Distribution   & Aug.  &  Boost                  & Network & +OMP    \\
\midrule
0  &  easy                   & no    &  no                     &  0.042  &  0.347  \\
0  &  hard                   & no    &  no                     &  0.126  &  0.653  \\
\midrule
0  &  hard                   & axis  &  no                     &  0.133  &  0.691  \\
0  &  hard                   & time  &  no                     &  0.138  &  0.736  \\
0  &  hard                   & both  &  no                     &  0.143  &  0.772  \\
0  &  hard                   & both  &  yes                    &  0.171  &  0.859  \\
\midrule
1  &  hard                   & both  & yes                     &  0.196  &  0.893  \\
2  &  hard                   & both  & yes                     &  0.217  &  0.925  \\
3  &  hard                   & both  & yes                     &  0.223  &  0.941  \\
\midrule
3  &  hard + ext.         & both  & yes                        &  0.283  &  1.00   \\
\bottomrule
\end{tabular}
\label{tab:ablation_study}
\end{table}

\begin{figure}[b]
    \centering
	\includegraphics[width=1.0\linewidth]{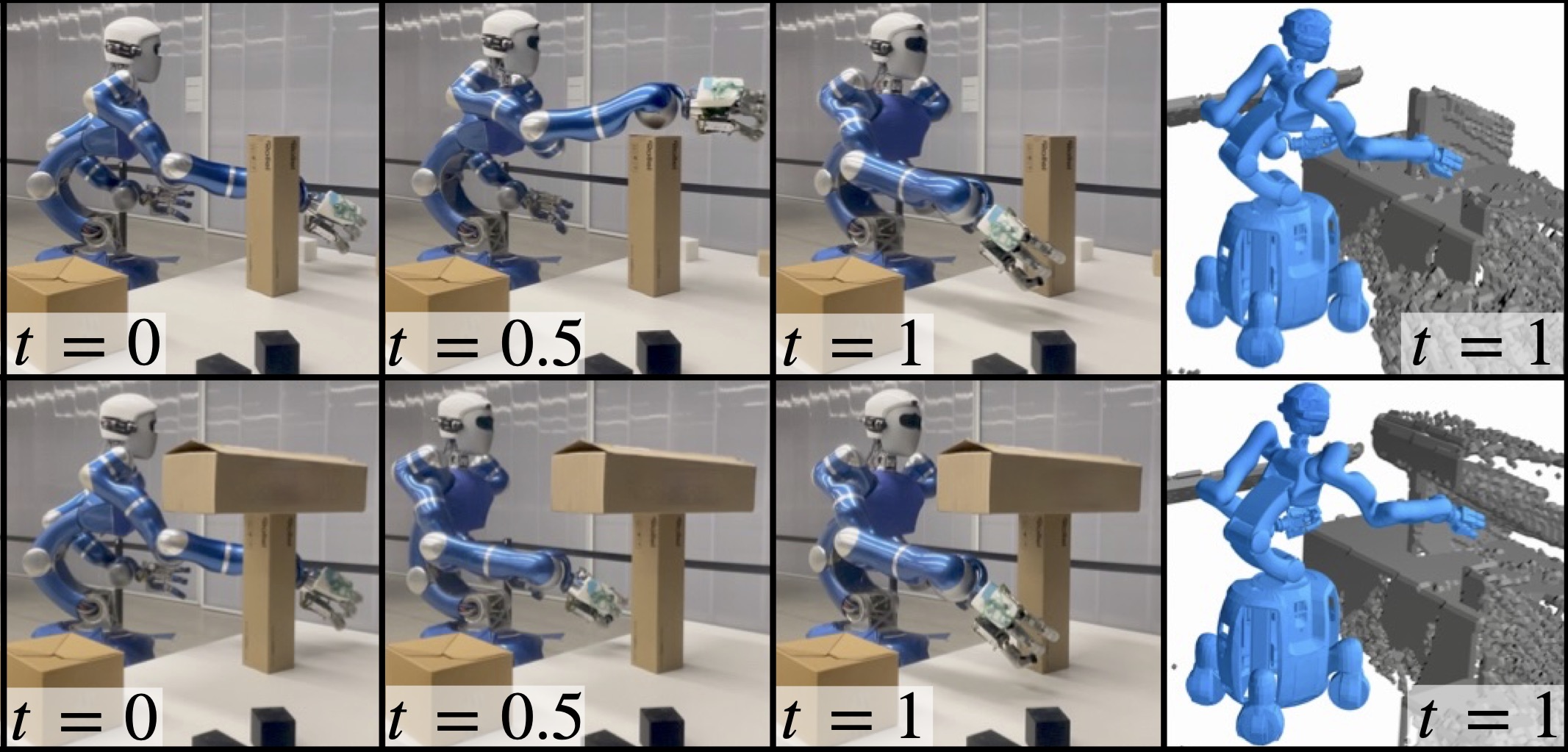}
	\vspace*{-5mm}
	\caption{
Agile Justin in two table scenes with boxes. 
The robot applies a different strategy for obstacle avoidance after the top route is blocked.
The rendered voxel model on the right shows the input for the neural network.}
	\label{fig:Justin_sim2real}
\end{figure}

\citet{Prokudin2019} demonstrated that the BPS with fully connected layers is superior to occupancy maps with CNNs or point clouds with a PointNet architecture, both in terms of required network parameters and training performance. 
We can confirm those findings for motion planning. 
The large memory requirements in 3D made training prohibitively slow and hard to iterate on network architecture or training methods.
Furthermore, looking towards the application on Agile Justin, the BPS can readily be integrated with the high-resolution SDFs acquired from the robot's depth camera~\cite{Wagner2013}.

The BPS representation and the proposed training scheme on the worlds from simplex noise were robust enough to even generalize to some first results on a real robot (see \cref{fig:Justin_sim2real}).
Only trained on those random worlds, the network was able to make valuable predictions from the data collected by Agile Justin's depth camera~\cite{Wagner2013}.
The predictions as warm-start for OMP could solve the unseen motion tasks which needed multi-starts otherwise in under 200\,ms.

Learning-based motion planning for such a complex robot was not tackled before in unseen environments, so we only compare it against simpler problems.
MPNetPath, for example, takes $0.59\,s$ to plan for the 7\,DoF Baxter arm in a known table scene~\cite{Qureshi2021}.
Non-learning-based methods like CHOMP take multiple seconds for similar scenes~\cite{Schulman2014}.

\cref{fig:NetworkBehaviour} shows a qualitative analysis of the network predictions. 
In motion planning, small changes in the problem often lead to fundamentally different solutions.
Our worlds and training were challenging enough that the networks react sharply to small changes in the input, predicting completely different solutions to only slightly altered problems.

\section{Conclusions}\label{sec:Conclusions}

We successfully trained motion planning networks using supervised learning on diverse and challenging datasets that predict paths close to the global optimum for previously unseen environments.
Using this prediction as a warm-start for optimization-based motion planning massively outperforms random multi-start. 
For the complex robot Agile Justin with 19 DoF, planning takes only 200\,ms on a single CPU core. 
This shows for the first time that learning-based motion planning works in previously unseen environments for such a complex robot.

One key to success is the basis point set encoding for the environment borrowed from computer vision which we introduced to motion planning and scales well to high-resolution 3D worlds.
In addition, we autogenerate a training dataset of hard examples, i.e., situations for which the vanilla OMP struggles and for which the trained network should later provide an educated initial guess.
We also introduced a scheme to further adapt the dataset during training by cleaning, boosting, and extending the dataset based on a metric defined by the (current) neural network and the objective function of the OMP.
This approach leads to a challenging and consistent dataset on which a network can efficiently be trained and improved.

In the future, we extend the planning problem towards manipulation and grasping by incorporating the inverse kinematics so that no longer a goal configuration but only the goal pose of the end-effector has to be provided. 
We will also further investigate and increase the real-world capabilities of our method.
As it is expensive to create a vast amount of real-world data, the goal is that our architecture and the autogenerated dataset allow for a robust transfer of the experience to real scenes.

\footnotesize
\bibliographystyle{IEEEtranN-modified}
\bibliography{IEEEabrv, references.bib}
\end{document}